%% file: main.tex
\documentclass[runningheads]{llncs}
\PassOptionsToPackage{table,dvipsnames}{xcolor}

\usepackage{eccv}

\usepackage{eccvabbrv}

\usepackage{graphicx}
\usepackage{booktabs}

\usepackage{hyperref}

\usepackage{multirow}

\begin{document}

\title{Decomposed Vision-Language Alignment for Fine-Grained Open-Vocabulary Segmentation} 

\titlerunning{Decomposed VL Alignment for Fine-Grained OV Segmentation}

\author{Chenhao Wang\inst{1,2} \and
Yingrui Ji\inst{1,2} \and
Yu Meng\inst{1} \and
Yao Zhu\inst{3}}


\institute{Aerospace Information Research Institute, Chinese Academy of Sciences \and
University of Chinese Academy of Sciences \qquad \textsuperscript{3}Zhejiang University \\
\email{wangchenhao22@mails.ucas.ac.cn, jiyingrui1996@gmail.com, mengyu@aircas.ac.cn, ee\_zhuy@zju.edu.cn}}

\maketitle

\begin{abstract}
Open-vocabulary segmentation models often struggle to generalize to unseen combinations of object categories and attributes, because fine-grained descriptions are typically encoded as holistic sentences that entangle multiple semantic units. We propose a Decomposed Vision-Language Alignment framework that explicitly factorizes textual prompts into a concept token and multiple attribute tokens, enabling separate cross-modal interactions for each semantic unit. At the feature level, we introduce a Feature-Gated Cross-Attention module that generates attribute-specific gating maps to fuse information in a multiplicative manner, effectively enforcing compositional semantics. At the scoring level, per-token similarities are aggregated in log-space, producing a stable and interpretable compositional matching. The method can be seamlessly integrated into existing transformer-based segmentation architectures and significantly improves generalization to unseen attribute–category compositions in fine-grained open-vocabulary segmentation benchmarks.
\end{abstract}

\input{sec/1.intro}
\input{sec/2.relatedwork}
\input{sec/3.method}

\input{sec/4.experiments}
\input{sec/5.conclusion}

\par\vfill\par


%
%
\bibliographystyle{splncs04}
\bibliography{main}
\end{document}

%% file: sec/1.intro.tex
\section{Introduction}

Open-vocabulary segmentation (OVS) aims to localize and segment image regions 
conditioned on arbitrary textual prompts. Benefiting from large-scale 
vision–language pretraining~\cite{radford2021learning}, OVS moves beyond 
conventional closed-set category vocabularies and enables models to generalize 
to previously unseen semantic concepts at inference 
time~\cite{li2022language,ghiasi2022openseg,xu2022groupvit,liang2023open,
xu2023san,xu2023ovsegmentor}. In real-world scenarios, targets are often 
described not only by object categories but also by multiple attributes, such 
as industrial building with flat roof or residential building with gable roof. 
These descriptions form compositional semantics, where a category is jointly 
specified by several attributes. Training data rarely covers all possible 
attribute–category combinations, so models must generalize to unseen 
compositions at test time.

Compositional generalization has been extensively studied in Compositional 
Zero-Shot Learning (CZSL). Prior works represent categories as attribute–object 
pairs and perform compatibility modeling~\cite{frome2013devise,
akata2016labelembedding}, graph-based propagation~\cite{naeem2021compcos}, or 
feature factorization~\cite{nagarajan2018attributes,chen2020compositionalfewshot} 
to recognize unseen compositions at the image level. In the generalized CZSL 
setting, bias calibration and harmonic mean evaluation protocols are introduced 
to balance seen and unseen compositions~\cite{xian2019zslcomprehensive}. These 
methods explicitly decompose semantic descriptions into attribute and object 
components and model their interactions in the label embedding space. 
Decomposition is performed at the classification level: each composition is 
scored by measuring compatibility between a global image representation and a 
composed label vector. This design is tailored for image-level recognition and 
does not carry over to dense prediction, where semantic units must be 
independently aligned with spatial regions rather than composed in a label space 
disconnected from pixel-level features.

Recent analyses further indicate that even large-scale vision–language models 
exhibit systematic biases in fine-grained compositional 
reasoning~\cite{yuksekgonul2023bow,thrush2022winoground}. In detection, 
localization, and instance segmentation, the challenge shifts from label-level 
composition to cross-modal alignment of semantic units at region and pixel 
levels. Although existing open-vocabulary detection and segmentation methods 
leverage vision–language pretraining to enhance category 
generalization~\cite{zareian2021ovdc,xu2022simplebaselineovss}, they encode the 
entire textual prompt as a single sequence. Such holistic encoding entangles 
category and attribute semantics in a shared representation. The compositional 
structure within the prompt is not explicitly modeled. When certain 
attribute–category combinations are absent during training, the model has no 
mechanism to decompose and recombine semantic units at inference, leading to 
degraded performance on unseen compositions. Transferring the decomposition 
insight from CZSL to this setting therefore requires a different formulation: 
each semantic unit must interact with visual features independently through 
cross-modal attention, and their outputs must be composed with explicit 
AND-style constraints at the feature and scoring levels.

To address this, we propose a decomposed vision–language compositional modeling 
framework for fine-grained open-vocabulary segmentation. Each textual prompt is 
decomposed into a category token and multiple attribute tokens, and cross-modal 
interaction is performed with each semantic unit separately, avoiding the 
entanglement introduced by holistic encoding. At the feature level, we introduce 
a Feature-Gated Cross-Attention module that generates attribute-specific gating 
maps and enforces \textsc{AND} constraints via element-wise multiplication, 
treating each attribute as a necessary condition filter. At the scoring level, 
we propose a log-space compositional aggregation strategy that sums per-unit 
matching scores in log-probability space, yielding a numerically stable 
formulation of \textsc{AND} semantics with improved optimization behavior.

We construct a compositional generalization evaluation protocol that explicitly 
includes unseen attribute–category compositions and conduct extensive experiments 
on two benchmarks.

Our main contributions can be summarized as follows:
\begin{itemize}
    \item We propose a Decomposed Vision-Language Alignment framework that can 
    be seamlessly integrated into existing transformer-based segmentation 
    architectures, significantly improving fine-grained open-vocabulary 
    segmentation.
    \item We propose Feature-Gated Cross-Attention, where attributes act as 
    necessary-condition filters, decoupling attribute-specific feature learning.
    \item We introduce a Log-Space \textsc{AND} Compositional Scoring scheme 
    that explicitly models logical \textsc{AND} relationships among semantic 
    units.
    \item We introduce a compositional generalization evaluation protocol for 
    open-vocabulary segmentation, and conduct extensive experiments that 
    demonstrate the effectiveness of our framework.
\end{itemize}

%% file: sec/2.relatedwork.tex
\section{Related Work}
\subsection{Open-Vocabulary Segmentation}
Traditional semantic segmentation models operate under a closed-set assumption, requiring retraining whenever new categories are introduced~\cite{long2015fully, chen2017deeplab, cheng2021per}. Open-vocabulary segmentation (OVS) relaxes this constraint by conditioning on free-form text prompts, enabling recognition of arbitrary categories at inference time. CLIP~\cite{radford2021learning} and ALIGN~\cite{jia2021scaling} establish joint image--text embedding spaces through contrastive pretraining, and have become the standard backbone that OVS methods build upon for dense prediction.

Early OVS approaches follow a two-stage pipeline: a class-agnostic proposal network generates mask candidates, which are then ranked using CLIP similarity. OpenSeg~\cite{ghiasi2022openseg}, ZSSeg~\cite{xu2021simple}, and ZegFormer~\cite{ding2022decoupling} are representative of this decoupled design. Subsequent methods tighten the integration between vision--language alignment and segmentation. OVSeg~\cite{liang2023open} fine-tunes CLIP on masked image regions to better handle partial observations; FC-CLIP~\cite{yu2024fcclip} processes both tasks within a single convolutional pass; CAT-Seg~\cite{cho2024cat} constructs explicit cost volumes between image and text features to improve localization precision. In remote sensing, MovSeg~\cite{ji2026movseg} extends this line by incorporating near-infrared bands alongside RGB through parameter-efficient adaptation.
A related direction focuses on sub-sentence grounding, aligning individual phrases with specific image regions. GLIP~\cite{li2022glip} reformulates detection as phrase grounding, enabling region--text correspondence below sentence level. MDETR~\cite{kamath2021mdetr} conditions cross-attention on different text spans to handle multi-phrase queries, and X-Decoder~\cite{zou2023xdecoder} unifies referring segmentation with panoptic segmentation under a shared text-conditioned query mechanism.

SAM~\cite{kirillov2023segment} establishes strong promptable segmentation capabilities by training on over one billion masks, and SAM~2~\cite{ravi2024sam2} extends this to video with memory-augmented decoding. Building upon these foundations, SAM~3~\cite{carion2025sam} further unifies geometric localization with native vision--language reasoning. Several works adopt this family of foundation models as frozen mask generators, pairing them with separate vision--language modules for semantic labeling. SOPSeg~\cite{wang2025sopseg} follows this paradigm for small object segmentation in remote sensing, combining region-adaptive magnification with an edge-aware decoder to preserve fine spatial details.
Referring image segmentation addresses a closely related problem of localizing regions described by natural language. LAVT~\cite{yang2022lavt} and CRIS~\cite{wang2022cris} align visual and linguistic features through cross-modal attention, while GRES~\cite{liu2023gres} handles more general expressions that may refer to multiple instances or no target at all.

\subsection{Compositional Zero-Shot Learning}

Compositional zero-shot learning (CZSL) studies the recognition of 
attribute--object pairs unseen during training, where both the attribute 
and object have been observed individually. 
Nagarajan and Grauman~\cite{nagarajan2018attributes} model attributes as 
linear operators that transform object embeddings, allowing unseen 
compositions to be synthesized algebraically. 
SymNet~\cite{li2020symmetry} imposes symmetry constraints to encourage 
consistent attribute transformations across different object categories. 
CGE~\cite{naeem2021cge} propagates semantic relationships over a learned 
graph to improve transfer to novel compositions. 
These methods share a common design: each composition is represented as a 
single global vector, and recognition is performed by measuring compatibility 
between that vector and a global image representation. 
Decomposition occurs in the label embedding space rather than in the feature 
space, so the resulting representations carry no spatial structure. 

This design is well suited to image classification but does not transfer 
directly to dense prediction tasks. 
Segmentation requires each semantic unit to be independently aligned with 
spatial regions at the pixel level, not composed in a label space that is 
disconnected from local visual features. 
A global composition vector has no mechanism to selectively activate on 
specific spatial locations corresponding to individual attribute conditions. 

Existing open-vocabulary segmentation methods do not address this gap either. 
They encode the full textual prompt as a single sequence, entangling category 
and attribute semantics in a shared representation~\cite{yuksekgonul2023when, 
doveh2023teaching}. 
When certain attribute--category combinations are absent from training, 
the model has no way to decompose and recombine semantic units at 
inference time.

%% file: sec/3.method.tex
\section{Method}

\subsection{Problem Formulation}

We study open-vocabulary fine-grained compositional segmentation under a compositional generalization setting. Unlike open-vocabulary segmentation, which predicts masks conditioned solely on category texts, our task requires the model to localize instances specified by a category together with multiple attributes, and to generalize to novel attribute combinations that are not observed during training.

Let $x \in \mathcal{X}$ denote an input image. Let $\mathcal{C}$ be the set of object categories, and let $\mathcal{A}_1, \mathcal{A}_2, \dots, \mathcal{A}_K$ denote $K$ attribute groups. A compositional semantic description is defined as
\begin{equation}
y = (c, a_1, a_2, \dots, a_K), 
\end{equation}
where $c \in \mathcal{C}$ and $a_k \in \mathcal{A}_k$. Given an image--text pair $(x, y)$, the goal is to predict all instance masks $\{ m_i \}_{i=1}^{N} \subset \mathcal{M}$ corresponding to image regions that simultaneously satisfy all specified semantic constraints in $y$. 
Formally, we learn a mapping 
\begin{equation}
f: (\mathcal{X}, \mathcal{Y}) \rightarrow \mathcal{P}(\mathcal{M}), 
\end{equation}
where $\mathcal{P}(\mathcal{M})$ denotes the set of mask subsets and $\mathcal{Y} \subseteq \mathcal{C} \times \mathcal{A}_1 \times \dots \times \mathcal{A}_K$ is the space of valid compositions.

The compositional generalization setting assumes that the training set contains only a subset of possible compositions, denoted as $\mathcal{Y}_{\text{seen}} \subset \mathcal{Y}$. The test set is constructed exclusively from held-out compositions 
$\mathcal{Y}_{\text{unseen}}$, where $\mathcal{Y}_{\text{seen}} \cap \mathcal{Y}_{\text{unseen}} = \emptyset$.

Importantly, each category $c$ and each individual attribute $a_k$ appear in the training data, but certain combinations of them are intentionally held out. At test time, the model is required to correctly segment instances corresponding to such unseen compositions.

This setting differs from standard open-vocabulary segmentation, which primarily focuses on generalizing to novel category names. In contrast, our problem emphasizes generalization over structured semantic compositions, where the key challenge lies in modeling the interaction among semantic components at the pixel level.

\subsection{Explicit Prompt Decomposition}

\begin{figure*}[htbp]
    \centering
    \includegraphics[width=\textwidth]{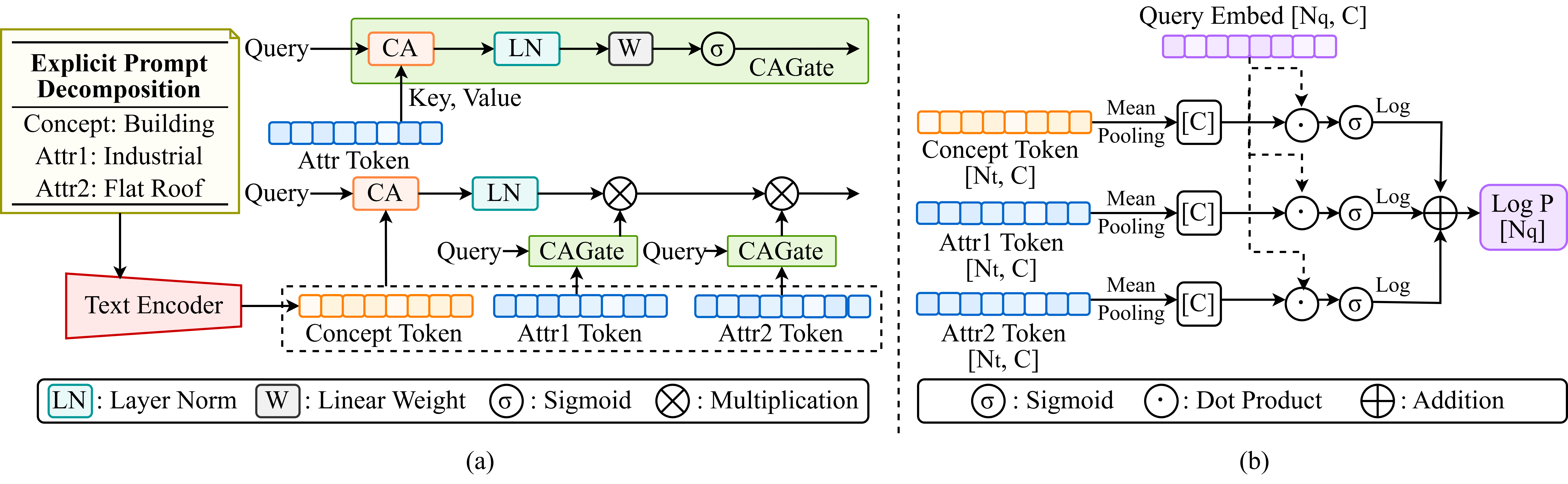}
    \caption{Overall architecture of the proposed method. (a) \textbf{Explicit Prompt Decomposition and Feature-Gated Cross-Attention}. The compositional text prompt is explicitly decoupled into independent concept and attribute tokens. Visual queries interact with the concept via standard cross-attention, and with attributes via multiplicative feature-gating to enforce compositional constraints. (b) \textbf{Log-Space AND Compositional Scoring}. Concept and attribute tokens are independently matched with query embeddings. The resulting scores are converted to log-probabilities and aggregated additively to produce the final compositional matching score.}
    \label{fig:architecture}
    
\end{figure*}

Encoding the concept and attributes as a single natural-language sentence 
typically leads vision--language models to learn highly entangled joint 
representations, which weakens generalization to unseen compositions. To 
mitigate this issue, as illustrated in Figure~\ref{fig:architecture}(a), we 
explicitly encode the category and each attribute independently while preserving 
the compositional structure defined in Sec.~3.1.

Given a compositional semantic description $y = (c, a_1, a_2, \dots, a_K)$, 
we encode the category and attributes separately rather than concatenating them 
into a single sentence. The category text $c$ is fed into the language encoder 
to obtain a sequence of token embeddings $t_c \in \mathbb{R}^{N_t \times C}$, 
where $N_t$ denotes the number of tokens for the category phrase and $C$ is the 
embedding dimension.

Attributes are encoded independently of each other, but not independently of 
the category. An attribute phrase such as flat roof is geometrically ambiguous 
when encoded in isolation: the same token may activate on any flat surface 
rather than specifically on building rooftops. To ensure each attribute token 
captures the intended object-conditioned meaning, we concatenate the category 
text with each attribute phrase before encoding. For example, flat roof is 
encoded as flat roof building, and residential is encoded as residential 
building. This conditioning anchors the attribute representation to the target 
object class while keeping different attributes disentangled from each other. 
Each conditioned attribute phrase is independently encoded as 
$t_{a_k} \in \mathbb{R}^{N_t \times C}$.

Since different samples may contain different numbers of attributes, we 
introduce an attribute presence mask $m \in \{0,1\}^{K}$, where $m_k = 1$ 
indicates that the $k$-th attribute is present. This mask ensures neutral 
behavior for missing attributes in the subsequent compositional aggregation. 
For each sample, the structured textual representation consists of the category 
token sequence $t_c$, the set of attribute token sequences $\{t_{a_k}\}_{k=1}^{K}$, 
and the presence mask $m$, providing a structured semantic interface for the 
feature-level and score-level compositional operations that follow.

\subsection{Decomposed Feature-Gated Cross-Attention}

As detailed in Figure~\ref{fig:architecture}(a), we perform decomposed cross-attention between the query (e.g., visual features or object queries) and each semantic unit independently, and enforce an explicit AND composition via feature-level multiplicative gating.

Let the query features be $\mathbf{Q} \in \mathbb{R}^{B \times N \times C}$, where $B$ is the batch size, $N$ the number of query tokens, and $C$ the feature dimension.
The category tokens are $\mathbf{T}_c \in \mathbb{R}^{B \times N_t \times C}$, and the $k$-th attribute tokens are $\mathbf{T}_{a_k} \in \mathbb{R}^{B \times N_t \times C}$.

\paragraph{\textbf{Category Interaction.}}
The category branch performs standard cross-attention:
\begin{equation}
\mathbf{Q}_c
=
\mathrm{LN}\!\left(
\mathrm{CA}(\mathbf{Q}, \mathbf{T}_c)
\right),
\end{equation}
where $\mathrm{CA}$ denotes multi-head cross-attention and $\mathrm{LN}$ denotes LayerNorm.

\paragraph{\textbf{Attribute Gates.}}
For each attribute $a_k$, we compute an independent gate:
\begin{equation}
\mathbf{G}_{a_k}
=
\mathrm{CAGate}(\mathbf{Q}, \mathbf{T}_{a_k})
=
\sigma \!\left(
W \cdot
\mathrm{LN}\!\left(
\mathrm{CA}(\mathbf{Q}, \mathbf{T}_{a_k})
\right)
\right),
\end{equation}
where $W$ is a learnable linear projection, $\sigma$ is the sigmoid function, and 
$\mathbf{G}_{a_k} \in [0,1]^{B \times N \times C}$.

Let $m_k \in \{0,1\}^{B}$ denote the attribute presence mask introduced in Sec.~3.2.
For samples where attribute $a_k$ is absent, the corresponding gate is neutralized:
\begin{equation}
\mathbf{G}_{a_k}
\leftarrow
m_k \cdot \mathbf{G}_{a_k}
+
(1 - m_k).
\end{equation}
which ensures the gate equals 1 when the attribute is absent.

\paragraph{\textbf{AND Composition.}}
All attribute gates are aggregated via element-wise multiplication:
\begin{equation}
\label{eq:and_comp}
\mathbf{G}_{\mathrm{total}}
=
\prod_{k=1}^{K}
\mathbf{G}_{a_k}.
\end{equation}

The final output feature is
\begin{equation}
\mathbf{F}_{\mathrm{out}}
=
\mathbf{Q}_c
\odot
\mathbf{G}_{\mathrm{total}},
\end{equation}
where $\odot$ denotes element-wise multiplication and $\alpha$ is a scaling factor. Each attribute gate can be interpreted as estimating the probability that the visual feature satisfies the corresponding attribute constraint. The multiplicative aggregation therefore approximates the logical AND over attributes, enforcing that a feature is preserved only when it is consistent with all attribute conditions.

\subsection{Log-Space AND Compositional Scoring}
\paragraph{\textbf{Decomposed Matching Scoring.}}

As shown in Figure~\ref{fig:architecture}(b), the category and attribute branches are fully decoupled during scoring. Let the decoder output query features be 
$\mathbf{H} \in \mathbb{R}^{B \times N \times C}$, where $B$ denotes the batch size, $N$ the number of queries, and $C$ the feature dimension.

For the category token sequence $\mathbf{T}_c$, we first perform mean pooling over valid tokens to obtain a category semantic vector. The pooled representation and the query features are then projected into a shared embedding space, followed by a scaled dot product to produce the category matching logits: $\mathbf{S}_c \in \mathbb{R}^{B \times N}$.

For each attribute $a_k$, similarly compute the corresponding matching logits: $\mathbf{S}_{a_k} \in \mathbb{R}^{B \times N}$.

The category and attribute branches are fully decoupled during scoring, with each semantic unit forming an independent matching branch.

\paragraph{\textbf{Log-Space AND Aggregation.}}

To model the semantic constraint of simultaneously satisfying the category and all attributes, we first map each branch logit to the probability of satisfying the corresponding semantic unit:
\begin{equation}
\pi_c = \sigma(S_c), 
\qquad
\pi_{a_k} = \sigma(S_{a_k}),
\end{equation}
where $\sigma(\cdot)$ denotes the sigmoid function.

In probability space, the AND semantics corresponds to multiplicative aggregation:
\begin{equation}
\pi_{\mathrm{AND}} 
= 
\pi_c \times \prod_{k} \pi_{a_k}.
\end{equation}

However, directly multiplying probabilities leads to rapid numerical decay and unstable gradients. We therefore perform an equivalent transformation in log-space. Taking the logarithm yields:
\begin{equation}
\log \pi_{\mathrm{AND}}
=
\log \pi_c
+
\sum_{k} \log \pi_{a_k}.
\end{equation}

In practice, we compute $\ell_c = \log \sigma(S_c)$, and $\ell_{a_k} = \log \sigma(S_{a_k}).$

Let $m_k \in \{0,1\}$ denote the attribute presence indicator. When an attribute is absent in a sample, its probability should be neutral (equal to 1), corresponding to 0 in log-space. Thus, $\ell_{a_k} \leftarrow m_k \cdot \ell_{a_k}$.

Finally, to maintain consistent scale across different numbers of active  attributes, we normalize by the number of effective components:
\begin{equation}
\ell
=
\frac{
\ell_c + \sum_{k} m_k \cdot \ell_{a_k}
}{
1 + \sum_{k} m_k
}.
\end{equation}

The resulting $\ell$ represents the log-probability that a query satisfies the compositional semantics, and is used as the final matching score.

\paragraph{\textbf{Log-space BCE.}}

Let $y \in \{0,1\}$ denote whether the current query matches the compositional semantic description.

The binary cross-entropy can be written as
\begin{equation}
\mathcal{L}
=
- y \cdot \ell
- (1 - y) \cdot \log \bigl(1 - \exp(\ell)\bigr),
\end{equation}
where $\ell$ is the aggregated log-probability defined before.

Compared with directly multiplying probabilities in the probability space, the log-space formulation avoids the exponential gradient attenuation caused by repeated products, leading to more stable optimization when multiple attributes are involved.

\subsection{Integration into SAM3 Framework}

Our method is fully model-agnostic: the proposed Feature-Gated Cross-Attention and log-space AND compositional scoring can be integrated with any open-vocabulary segmentation backbone, requiring only modular replacements in text encoding, cross-attention, and matching scoring.

In our experiments, we instantiate the method on SAM3 to leverage its strong open-vocabulary capabilities while keeping the detection and segmentation heads unchanged. The Feature-Gated Cross-Attention module is applied at multiple stages: 
(i) during image encoding, where visual features are semantically conditioned on decomposed textual units; 
(ii) during object query decoding, enforcing instance-level compositional constraints; 
(iii) during mask prediction, maintaining semantic consistency in the generated masks. 

Matching scores are computed via log-space AND aggregation, providing a differentiable supervisory signal for compositional semantics. This integration demonstrates that our compositional framework can endow any strong open-vocabulary backbone with fine-grained semantic generalization.

%% file: sec/4.experiments.tex
\section{Experiments}

\subsection{Compositional Generalization Evaluation Protocol}

To evaluate models’ ability to generalize to unseen attribute–category combinations under open-vocabulary conditions, we design a unified compositional generalization evaluation protocol that integrates dataset splitting, training label construction, and testing with comprehensive metrics.

\noindent\textbf{Dataset and Compositional Split.}
Experiments are conducted on two datasets: the UBC building dataset~\cite{Huang_2022_CVPR} and the PACO-LVIS dataset~\cite{Ramanathan_2023_CVPR}.
In the UBC dataset, each building instance is annotated with two independent attribute groups: roof type and building use.
To construct a compositional generalization setting, only four attribute combinations are retained in the training set as \emph{seen combinations}: (hipped roof, residential), (gable roof, commercial), (flat roof, industrial), and (other, public).
All remaining combinations are treated as \emph{unseen combinations} and removed from training.

To evaluate the method in more general visual scenarios, we further conduct experiments on the PACO-LVIS dataset.
PACO is an instance-level attribute dataset built upon LVIS, providing attribute annotations across four attribute groups: color, material, pattern/marking, and transparency.
To avoid ambiguity caused by multiple attributes of the same type, we select instances containing at most one attribute per attribute group.
From all possible pairs of attributes, we randomly sample one tenth as \emph{seen compositions}, while the remaining combinations are treated as unseen.

\noindent\textbf{Training Label Construction.}
To prevent the model from relying on a single prompt granularity, training labels are constructed at three levels of semantic specificity.
First, one third of the training images are annotated using the category label only (e.g., ``building'').
Second, one third use single-attribute prompts combined with the category (e.g., ``flat roof building'' or ``residential building'').
Third, for the remaining images, single-attribute prompts are also used, and if the ground-truth attribute pair belongs to the seen set, an additional annotation containing both attributes is provided (e.g., ``flat roof industrial building'').
This design exposes the model to atomic semantics while introducing only a small number of attribute compositions, preventing direct memorization of unseen combinations.

\noindent\textbf{Test Protocol.}
Based on the predefined seen compositions for each dataset, we construct two evaluation splits: a \emph{seen-composition} test set containing only combinations observed during training, and an \emph{unseen-composition} test set containing the remaining combinations.
\noindent\textbf{Evaluation Metrics.}
We report three complementary metrics to evaluate compositional generalization.
(i) \textbf{Compositional Generalization}: mask average precision (AP) on \emph{unseen-composition}, which directly measures the model’s ability to localize and segment unseen attribute compositions.
(ii) \textbf{Seen–Unseen Gap}: we report the relative performance drop
$R_d = 1 - \frac{\text{AP}_{\text{unseen}}}{\text{AP}_{\text{seen}} + \epsilon}$,
which quantify how much performance degrades when moving from seen to unseen compositions.
(iii) \textbf{AND Efficiency}: 
$\text{AND-Eff} = 
\frac{\text{AP}_{\text{unseen}}}
{\min_{g \in \mathcal{G}} (\text{AP}_g)}$,
where $\mathcal{G}$ denotes the set of attribute groups (e.g., roof type and building use in UBC, or color, material, pattern, and transparency in PACO).
This metric measures whether independently learned attribute semantics can be correctly combined at inference time. Values close to 1 indicate effective compositional reasoning, while significantly lower values suggest failure to bind attributes properly.

\subsection{Comparison with Baselines}

We fine-tuned all baseline models on the compositional split. This ensures a fair comparison. All methods used the identical training set and three-level label construction strategy. The baseline models kept their original text encoding architectures. They processed the training text as a single continuous sentence. Our method differs from the baselines only in the text encoding strategy and the multi-modal interaction mechanism.

Table~\ref{tab:comparison_ovs} compares our method with representative open-vocabulary segmentation approaches under the proposed compositional generalization protocol.

A consistent trend across all baselines is the large gap between seen and unseen compositions. On UBC, unseen AP ranges from 1.7 to 3.5 while AND-Eff stays below 0.51, indicating that these models largely rely on memorized attribute--category co-occurrences rather than learning semantics that generalize to novel combinations. The comparison between CAT-Seg and X-Decoder further illustrates this point. Despite achieving higher AP$_\text{seen}$ through its query-based decoding architecture, X-Decoder's AND-Eff drops to 0.493, below CAT-Seg's 0.507, suggesting that stronger category-level localization does not translate to better attribute binding. SAM3 with whole-prompt encoding raises AP$_\text{seen}$ to 15.0 on UBC, yet unseen AP remains at 4.0 with $R_d{=}0.733$, confirming that representational capacity alone cannot compensate for entangled semantic encoding.

On PACO, the absolute seen--unseen gap is smaller across all methods, with $R_d$ values well below those on UBC. However, AND-Eff remains below 0.80 for all baselines, and SAM3 reaches only 0.874, indicating that the compositional binding problem persists even when attribute distributions are more favorable.

Our method improves over SAM3 on both datasets. On UBC, unseen AP increases from 4.0 to 7.1 and AND-Eff rises from 0.533 to 0.959, approaching the regime where compositional performance matches marginal attribute capability. On PACO, the gains are more modest in absolute terms---unseen AP increases from 37.9 to 39.2 and AND-Eff from 0.874 to 0.907---consistent with the observation that baselines already generalize better on this split. The fact that AP$_\text{seen}$ also improves on both datasets rules out the possibility that the gains on unseen compositions come at the cost of seen-composition performance.

\begin{table*}[t]
\centering
\caption{Comparison with open-vocabulary segmentation methods under the compositional generalization protocol. $R_d$ denotes the relative performance drop from seen to unseen compositions. AND-Eff measures how effectively the model composes independently learned attribute semantics at inference.}
\label{tab:comparison_ovs}

\resizebox{\textwidth}{!}{
\begin{tabular}{l|c|cccc|cccc}
\toprule
\multirow{2}{*}{Method} & \multirow{2}{*}{Backbone}
& \multicolumn{4}{c|}{UBC}
& \multicolumn{4}{c}{PACO} \\

\cmidrule(lr){3-6} \cmidrule(lr){7-10}

& & AP$_{\text{unseen}}$ & AP$_{\text{seen}}$ & $R_d\downarrow$ & AND-Eff$\uparrow$
  & AP$_{\text{unseen}}$ & AP$_{\text{seen}}$ & $R_d\downarrow$ & AND-Eff$\uparrow$ \\

\midrule
OVSeg~\cite{liang2023open} & CLIP ViT-L
& 1.7 & 9.4 & 0.819 & 0.386
& 26.3 & 31.4 & 0.162 & 0.728 \\

FC-CLIP~\cite{yu2024fcclip} & ConvNeXt-L
& 2.4 & 10.9 & 0.780 & 0.453
& 29.7 & 34.8 & 0.147 & 0.763 \\

CAT-Seg~\cite{cho2024cat} & CLIP ViT-B
& 3.4 & 12.3 & 0.724 & 0.507
& 32.8 & 38.6 & 0.150 & 0.784 \\

X-Decoder~\cite{zou2023xdecoder} & Focal-L
& 3.5 & 13.6 & 0.743 & 0.493
& 33.6 & 39.5 & 0.149 & 0.792 \\

\midrule
SAM3~\cite{carion2025sam} & SAM3
& 4.0 & 15.0 & 0.733 & 0.533
& 37.9 & 42.2 & 0.102 & 0.874 \\

\rowcolor{gray!10}
\textbf{Ours} & SAM3
& \textbf{7.1} & \textbf{16.8} & \textbf{0.577} & \textbf{0.959}
& \textbf{39.2} & \textbf{43.2} & \textbf{0.093} & \textbf{0.907} \\

\bottomrule
\end{tabular}
}
\end{table*}

\subsection{Effect of Prompt Encoding Strategy}
\label{sec:prompt_encoding}

Table \ref{tab:prompt_strategy} compares different text encoding strategies under the same SAM3 backbone. The concatenated sentence baseline achieves an unseen AP of 4.0. Replacing it with a template ensemble strategy, which diversifies the surface form of input prompts, brings only a marginal improvement to 4.2. This result indicates that reformulating the prompt at the lexical level does not resolve the underlying issue: category and attribute semantics remain entangled in a shared embedding representation.

Encoding the category and each attribute as separate text sequences, then averaging their matching scores, raises the unseen AP to 5.3 and AND-Eff to 0.707. Explicit decomposition at the encoding stage clearly benefits compositional generalization. However, averaging scores across semantic units introduces a permissive aggregation behavior. A query that strongly matches the category but fails to satisfy an attribute constraint can still receive a moderately high average score, which weakens discrimination in fine-grained settings.

Replacing mean aggregation with log-space AND scoring addresses this issue by requiring all semantic units to be jointly satisfied. The unseen AP further increases to 7.1 and AND-Eff reaches 0.959. The gap between independent encoding with mean aggregation and our full method isolates the contribution of the AND composition mechanism, showing that the scoring strategy is as critical as the decomposition itself.

\begin{table}[t]
\centering
\setlength{\tabcolsep}{7pt}
\caption{Comparison of prompt encoding strategies under the same SAM3 backbone on UBC dataset. All variants share the same visual encoder and segmentation head; only the text encoding and matching pipeline differ.}
\label{tab:prompt_strategy}
\begin{tabular}{l|cc|c|c}
\toprule
Prompt Strategy & AP$_\text{seen}$ & AP$_\text{unseen}$ & $R_d\downarrow$ & AND-Eff$\uparrow$ \\
\midrule
Concatenated sentence      & 15.0 & 4.0 & 0.733 & 0.533 \\
Template ensemble ($n$=5)  & 15.3 & 4.2 & 0.725 & 0.560 \\
Independent + Mean score   & 15.4 & 5.3 & 0.656 & 0.707 \\
\midrule
\rowcolor{gray!10}
\textbf{Ours} (Decomposed + AND)  & \textbf{16.8} & \textbf{7.1} & \textbf{0.577} & \textbf{0.959} \\
\bottomrule
\end{tabular}
\end{table}

\subsection{Ablation on Aggregation Mechanism}

Section~\ref{sec:prompt_encoding} fixes the aggregation strategy to isolate the 
effect of encoding. Here we fix the encoding to the decomposed form and vary 
only the aggregation mechanism, treating these two as orthogonal design axes.

Table~\ref{tab:aggregation} compares different aggregation mechanisms under the 
same decomposed prompt encoding. Additive fusion achieves a seen AP of 16.1 but 
an unseen AP of only 5.1. When attribute features are summed, a strong response 
on one attribute can compensate for a weak response on another, making it 
difficult to enforce strict compositional constraints. Max-score aggregation 
shows a similar pattern: the seen AP rises to 16.5, but the unseen AP drops to 
4.3. Selecting the highest-scoring branch implements OR semantics, which is 
insensitive to whether all attributes are jointly satisfied.

Directly multiplying branch probabilities enforces AND semantics and improves 
the unseen AP to 5.7. However, the seen AP decreases to 15.6. Repeatedly 
multiplying sigmoid outputs causes the gradient magnitude to diminish rapidly 
with the number of attributes, making optimization unstable and leading to 
underfitting even on seen compositions. Adding feature-level gating before 
probability multiplication partially alleviates this issue. By concentrating 
per-branch distributions through spatial pre-filtering, the unseen AP increases 
to 6.2. The gradient problem in probability space nonetheless persists. 
Replacing the probability product with log-space AND aggregation resolves this 
by converting multiplicative operations into additions, which maintains stable 
gradient flow. The final model achieves a seen AP of 16.8 and an unseen AP of 
7.1, with the improvement over the feature gating plus probability product 
variant attributable to the log-space formulation.

\begin{table}[t]
\centering
\setlength{\tabcolsep}{5pt}
\caption{Comparison of compositional aggregation mechanisms on UBC dataset. All variants use the same decomposed prompt encoding and SAM3 backbone. Only the feature-level fusion and score-level aggregation differ.}
\label{tab:aggregation}
\begin{tabular}{l|cc|c|c}
\toprule
Aggregation Mechanism & AP$_\text{seen}$ & AP$_\text{unseen}$ & $R_d\downarrow$ & AND-Eff$\uparrow$ \\
\midrule
Additive fusion                     & 16.1 & 5.1 & 0.683 & 0.680 \\
Prob-space product (Eq.~(9))          & 15.6 & 5.7 & 0.635 & 0.760 \\
Max-score (OR semantics)            & 16.5 & 4.3 & 0.739 & 0.573 \\
\midrule
Feature gating + Prob-space product & 16.4 & 6.2 & 0.620 & 0.827 \\
\rowcolor{gray!10}
Feature gating + Log-space AND (\textbf{Ours}) & \textbf{16.8} & \textbf{7.1} & \textbf{0.577} & \textbf{0.959} \\
\bottomrule
\end{tabular}
\end{table}

\subsection{Comparison with Adapted CZSL Methods}

The ablations above evaluate design choices within our framework. We further 
compare against compositional zero-shot learning methods to assess whether 
existing composition strategies from image-level recognition transfer 
effectively to dense prediction. Table~\ref{tab:czsl_adapted} presents results 
for methods adapted into the SAM3 framework, with the visual encoder and 
segmentation head kept unchanged.

The AttrAsOp~\cite{nagarajan2018attributes} approach models attributes as linear 
transformations on category query embeddings. This design improves the unseen AP 
from 4.0 to 4.6. The improvement is limited by its spatial uniformity. A global 
linear operator cannot reflect that different attributes constrain distinct 
spatial regions of a target object. The CGE~\cite{naeem2021compcos} approach 
propagates relationships between attributes and objects over a learned graph. 
This relational modeling increases the unseen AP to 5.1. Graph convolution 
relies on rich node connectivity, and the small number of attribute and category 
types in this dataset causes over-smoothing during propagation.

Both AttrAsOp and CGE generate a single global composition vector and evaluate 
it against visual queries through a standard dot product. This matching mechanism 
lacks spatial selectivity. Our Feature-Gated Cross-Attention generates spatially 
varying gate maps for each individual attribute. Both our method and CGE utilize 
decomposed attribute modeling, but our method achieves an unseen AP of 7.1 
compared to 5.1 for CGE. This gap is directly driven by pixel-level spatial 
gating rather than global composition vectors.

\begin{table}[t]
\centering
\setlength{\tabcolsep}{5pt}
\caption{Comparison with compositional zero-shot learning methods adapted for dense prediction on UBC dataset. Each method is integrated into the SAM3 framework by replacing the composition module while keeping the visual encoder and segmentation head unchanged.}
\label{tab:czsl_adapted}
\begin{tabular}{l|cc|c|c}
\toprule
Method & AP$_\text{seen}$ & AP$_\text{unseen}$ & $R_d\downarrow$ & AND-Eff$\uparrow$ \\
\midrule
SAM3 (whole prompt)                                    & 15.0 & 4.0 & 0.733 & 0.533 \\
\midrule
AttrAsOp~\cite{nagarajan2018attributes} (adapted)     & 15.2 & 4.6 & 0.697 & 0.613 \\
CGE~\cite{naeem2021cge} (adapted)                     & 15.5 & 5.1 & 0.671 & 0.680 \\
\midrule
\rowcolor{gray!10}
\textbf{Ours} (Decomposed + AND)                      & \textbf{16.8} & \textbf{7.1} & \textbf{0.577} & \textbf{0.959} \\
\bottomrule
\end{tabular}
\end{table}


\subsection{Qualitative Analysis}
\begin{figure}[htbp] 
 \vspace{-1em}
    \centering 
    \includegraphics[width=0.99\textwidth]{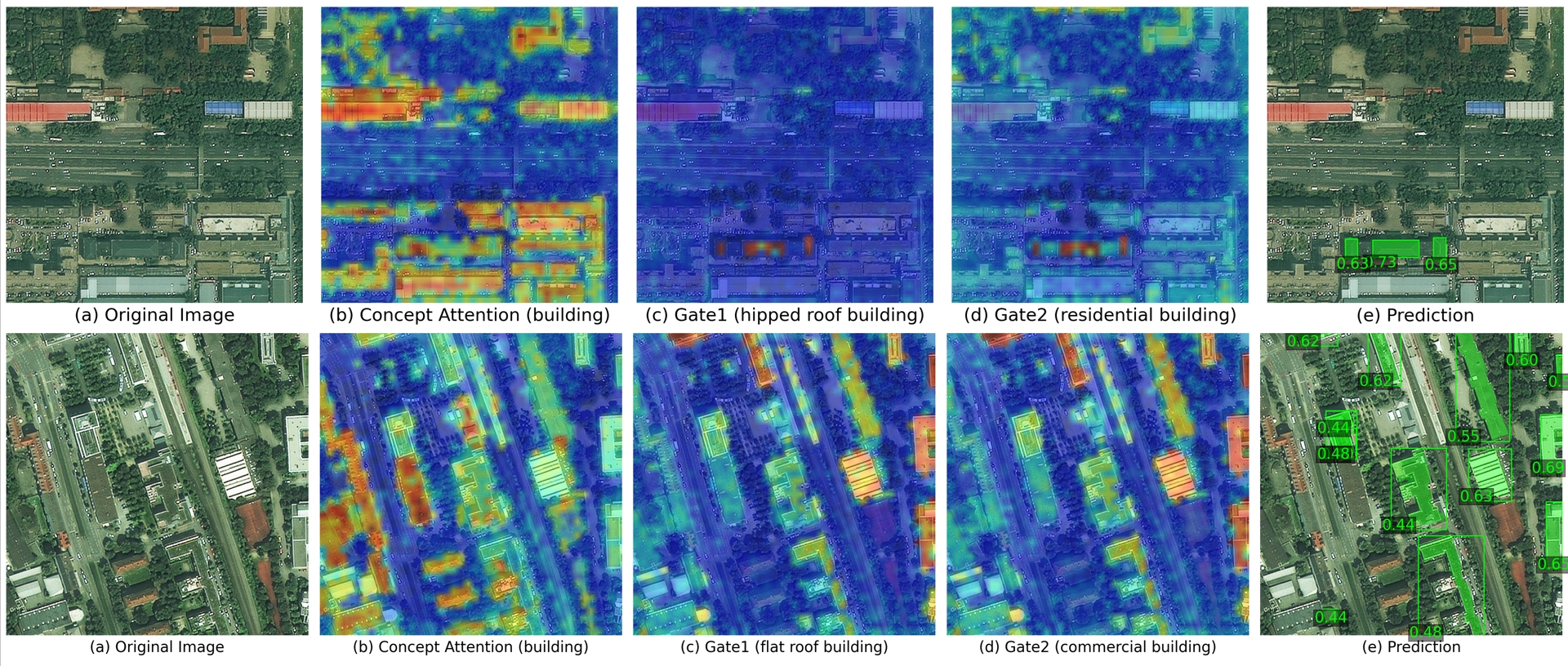} 
    \caption{Qualitative visualization of the Feature-Gated Cross-Attention mechanism for two examples. The top row prompt decomposes into concept building, attribute hipped roof building, and attribute residential building. The bottom row prompt decomposes into concept building, attribute flat roof building, and attribute commercial building. Panel a shows the original images. Panel b shows the concept attention maps. Panel c and Panel d display the CAGate activation maps for the two attributes. Panel e shows the final segmentation results. The three branches produce spatially distinct activations. The segmented instances coincide with their intersection.}
    \label{fig:atten} 
    \vspace{-2em}
\end{figure}

Figure \ref{fig:atten} visualizes the intermediate feature maps of Feature-Gated Cross-Attention for two representative examples. The input prompts are decomposed into one concept token and two attribute tokens. The top row uses the concept token building. Its attribute tokens are hipped roof building and residential building. The bottom row also uses the concept token building. Its attribute tokens are flat roof building and commercial building.

Panel b in Figure \ref{fig:atten} displays the concept attention maps. These maps broadly activate across various building instances. They ignore specific roof types or building usages. They serve as category-level localizers.

The attribute gates exhibit distinct spatial patterns. Panel c shows the activations for the first attribute gate. The top row gate responds selectively to hipped roof structures. It suppresses flat-roofed buildings. The bottom row gate highlights flat roof buildings. Panel d shows the activations for the second attribute gate. The top row gate highlights residential buildings. It attenuates commercial and industrial instances. The bottom row gate focuses on commercial buildings.

Panel e displays the final predictions. The model retains only instances located in the intersection of all three activation regions. It suppresses buildings matching the concept but violating either attribute condition. This suppression occurs through the multiplicative gating in Equation \ref{eq:and_comp}.

%% file: sec/5.conclusion.tex
\section{Conclusion}

This work investigates the challenge of compositional generalization in fine-grained open-vocabulary segmentation, where models must recognize and localize objects described by novel combinations of categories and attributes. We propose a decomposed vision--language alignment framework that factorizes textual prompts into category and attribute tokens, enabling separate cross-modal interactions for each semantic unit. This enables the model to learn attribute-aware representations and recombine novel semantic concepts beyond those observed during training. Our approach includes a Feature-Gated Cross-Attention module that enforces compositional \textsc{AND} constraints and a log-space scoring strategy for stable, interpretable aggregation of per-token similarities. Extensive experiments under a compositional generalization protocol show that our framework significantly improves performance on unseen attribute–category combinations while maintaining strong results on seen ones. We hope this work motivates further research on compositional reasoning in vision–language models and inspires the development of more robust open-world perception systems.